\documentclass[10pt,twocolumn,letterpaper]{article}

\usepackage{cvpr}
\usepackage{times}
\usepackage{epsfig}
\usepackage{graphicx}
\usepackage{amsmath}
\usepackage{amssymb}

\usepackage{array}
\usepackage{tabularx}
\newcolumntype{L}[1]{>{\raggedright\arraybackslash}p{#1}}
\newcolumntype{C}[1]{>{\centering\arraybackslash}p{#1}}
\newcolumntype{R}[1]{>{\raggedleft\arraybackslash}p{#1}}

\usepackage{times}  
\usepackage{helvet}  
\usepackage{courier}  
\usepackage{url}  
\usepackage{graphicx}  
\usepackage{booktabs}
\usepackage{appendix} 
\usepackage{pifont}

\usepackage{xcolor}
\usepackage{soul}
\usepackage[utf8]{inputenc}
\usepackage[small]{caption}
\usepackage{verbatim}
\usepackage{amsfonts}
\usepackage{mathtools}

\usepackage[ruled,vlined]{algorithm2e}
\usepackage{breqn}
\usepackage{multirow}

\usepackage{times}
\usepackage{gensymb}
\usepackage{helvet}
\usepackage{courier}
\usepackage{url}

\usepackage[breaklinks=true,bookmarks=false]{hyperref}

\cvprfinalcopy 

\begin{document}

\title{3D Shape Reconstruction from a Single 2D Image via 2D-3D Self-Consistency}

\author{Yi-Lun Liao\thanks{Indicates equal contribution.}, Yao-Cheng Yang${^\ast}$, Yu-Chiang Frank Wang\\
Department of Electrical Engineering, National Taiwan University\\
{\tt\small \{b03901001, b03901161, ycwang\}@ntu.edu.tw}
}

\maketitle

\maketitle
\begin{abstract}

Aiming at inferring 3D shapes from 2D images, 3D shape reconstruction has drawn huge attention from researchers in computer vision and deep learning communities.
However, it is not practical to assume that 2D input images and their associated ground truth 3D shapes are always available during training. In this paper, we propose a framework for semi-supervised 3D reconstruction. This is realized by our introduced 2D-3D self-consistency, which aligns the predicted 3D models and the projected 2D foreground segmentation masks. Moreover, our model not only enables recovering 3D shapes with the corresponding 2D masks, camera pose information can be jointly disentangled and predicted, even such supervision is never available during training. In the experiments, we qualitatively and quantitatively demonstrate the effectiveness of our model, which performs favorably against state-of-the-art approaches in either supervised or semi-supervised settings.

\end{abstract}
\section{Introduction}

3D modeling and reconstruction can be applied to a variety of real-world applications, including visual rendering, modeling, and robotics. 
While it might not be difficult for human to infer 3D information from the observed 2D visual data, it is a very challenging task for machines to do so. 
Without sufficient 3D shapes, viewpoint information or 2D images from different viewpoints for as training data, it is difficult to reconstruct 3D models using 2D data.

Over the past few years, Convolution Neural Network (CNN) and generative adversarial networks (GAN)~\cite{GAN} have shown impressive progresses and results particularly in the areas of computer vision and image processing. For 3D shape reconstruction, several solutions have been proposed~\cite{TL,3DR2N2,sketch1,sketch2,dolphin,ray1,unsup1,NIPS2016_6206,multiple_obj,Weakly,semantic,MVDepthMap, Marrnet,MVDepthMap,lsmKarHM2017,mvc18,2Ddeconv,rethinking,pointset,visual_hull}.
As discussed later in Section~\ref{sect2}, different settings and limitations such as the required number of 2D input images, availability of the viewpoint information, and supervision of ground truth labels would limit the use of existing models for 3D reconstruction applications.

In this paper, we address a challenging task of 3D model reconstruction from a single 2D image. That is, during training and testing, we only allow our model to observe a single 2D image as the input, without knowing its viewpoint information (e.g., azimuth, elevation, and roll). If full label supervision is available, we allow both 3D ground truth shape (in voxels) and the associated 2D foreground segmentation mask to guide the learning of our proposed model. As for unlabeled data during the semi-supervised learning process, we only observe 2D images for training our model.

To handle the aforementioned challenging setting, we propose a deep learning architecture which not only recovers 3D shape and 2D mask outputs with full supervision. We further exploit their 2D-3D self-consistency during the learning process, which is the reason why we are able to utilize unlabeled image inputs to realize semi-supervised learning. Finally, as a special characteristics of our model, we are able to jointly disentangle camera pose information during the above process, even no ground truth pose information is observed any time during training.

The contributions of this work are highlighted below:

\begin{itemize}
\item In this paper, a unique semi-supervised deep learning framework is proposed for 3D shape prediction from a single 2D image.
\item The presented network is trained in an end-to-end fashion, while 2D-3D self-consistency is particularly introduced to handle unlabeled training image data.
\item In addition to 3D shape prediction, our model is able to disentangle camera pose information from the derived latent feature space, while supervision of such information is never available during training.
\item Experimental results quantitatively and qualitatively show that our method performs favorably against state-of-the-art fully supervised and weakly-supervised methods.

\end{itemize}

\section{Related Work}
\label{sect2}

\subsection{Learning for 3D Model Prediction}

Most existing methods for 3D shape reconstruction require supervised settings, i.e., both 2D input images and their corresponding 3D models need to be observed during the training stage. With the development of large-scale shape repository like ShapeNet~\cite{ShapeNet}, several methods of this category have been proposed~\cite{TL,3DR2N2}. For example, the work of Girdhar \textit{et al.}~\cite{TL} is fully trained with pairwise 2D images and 3D models, which is realized by learning joint embedding for both 3D shapes and 2D images. On the other hand, Wang \textit{et al.}~\cite{visual_hull} choose to predict 3D shape, 2D mask, and pose simultaneously, followed by construction of probabilistic visual hulls. However, they require additional ground truth pose information to train their model.

Since it might not be practical to assume that the ground truth 3D object information is always available, recent approaches like~\cite{NIPS2016_6206,Weakly,Marrnet,MVDepthMap,mvc18} manage to learn object representations in weakly-supervised settings.
For example, Yan \textit{et al.}~\cite{NIPS2016_6206} present Perspective Transformer Nets.
Guided by the projected 2D masks and given camera viewpoints, their proposed network architecture learns the perspective transformations of the target 3D object.
As a result, the 3D objects can be recovered using 2D images without the supervision of ground truth shape information.
Alternatively, Gwak \textit{et al.}~\cite{Weakly} take 2D images and viewpoint information as input, and utilize 2D masks as weak supervision information, which is enabled by perspective projection of reconstructed 3D shapes to foreground masks.
They constrain the reconstructed 3D shapes to a manifold observed by unlabeled shapes of real-world objects. Soltani \textit{et al.}~\cite{MVDepthMap} learn a generative model over depth maps or their corresponding silhouettes, and then use a deterministic rendering function to construct 3D shapes from the generated depth maps and silhouettes.
Although their network is able to generate images at different viewpoints from a single input when testing, it requires both silhouettes and depth maps from different views as ground truth for training purposes. To reconstruct 3D shapes, Tulsiani \textit{et al.}~\cite{mvc18} choose to utilize the consistency between shape and viewpoint information independently predicted from two camera views of the same instance.
However, they also require multiple images of the same instance taken from different camera viewpoints for training purposes.

Different from the works requiring images taken by different cameras for 3D reconstruction, our method only needs a single image for 3D shape reconstruction, and can be trained in a semi-supervised setting. Moreover, as noted and discussed in the following subsection, our model is able to disentangle the camera pose without the associated ground truth information.

\subsection{Learning Interpretable Representation}
Learning interpretable representation, or representation disentanglement, has attracted the attention fro researchers in the fields of computer vision and machine learning. Although Kingma \textit{et al.}~\cite{VAE} utilize variational autoencoders to handle the generality of objects with input noise for improved data distribution.
However, there is no guarantee that particular attributes in the derived latent space would correspond to desirable features. When it comes to recover 3D information from input 2D images, it is often desirable to be able to separate external parameters like viewpoint during the learning of visual object representation, so that the output data can be properly recovered or manipulated.

For representation disentanglement in 3D shape reconstruction, some existing approaches such as Adversarial Autoencoders (AAE)~\cite{AAE} manage to derive disentangled representations by supervised learning, matching the derived representations with specific labels and adversarial losses calculated from the discriminators. Weakly-supervised learning methods have also been proposed, which alleviate the need of utilizing fully labeled training data in the above process. For example, Deep Convolutional Inverse Graphics Network (DC-IGN)~\cite{DCIGN} clamps certain dimensions of representation vectors from a mini-batch of training instances for retrieving factors such as azimuth angle, elevation angle, azimuth of light source, or intrinsic properties. To perform representation disentanglement in a unsupervised setting, Information Maximizing GAN (InfoGAN)~\cite{InfoGAN} chooses to maximize mutual information between latent variables and observation to learn disentangled representations.

Recently, Grant \textit{et al.}~\cite{DeepDisentangled} propose Deep Disentangled Representations for Volumetric Reconstruction, which takes 2D images as inputs and produces separate representations for 3D shapes of objects and parameters of viewpoint and lighting. Since their approach requires full data supervision for deriving the associated shape and transformation information (i.e., ground truth volumetric shapes always available during training), it would not be easy to extend their work to practical scenarios where only a portion of 2D image data are with ground truth 3D information. Sharing similar goals, we aim at identifying shape and viewpoint information in a semi-supervised setting, while no ground truth camera pose information is observed during training.
\begin{figure*}[t]
  \centering	  \includegraphics[width=0.875\textwidth]{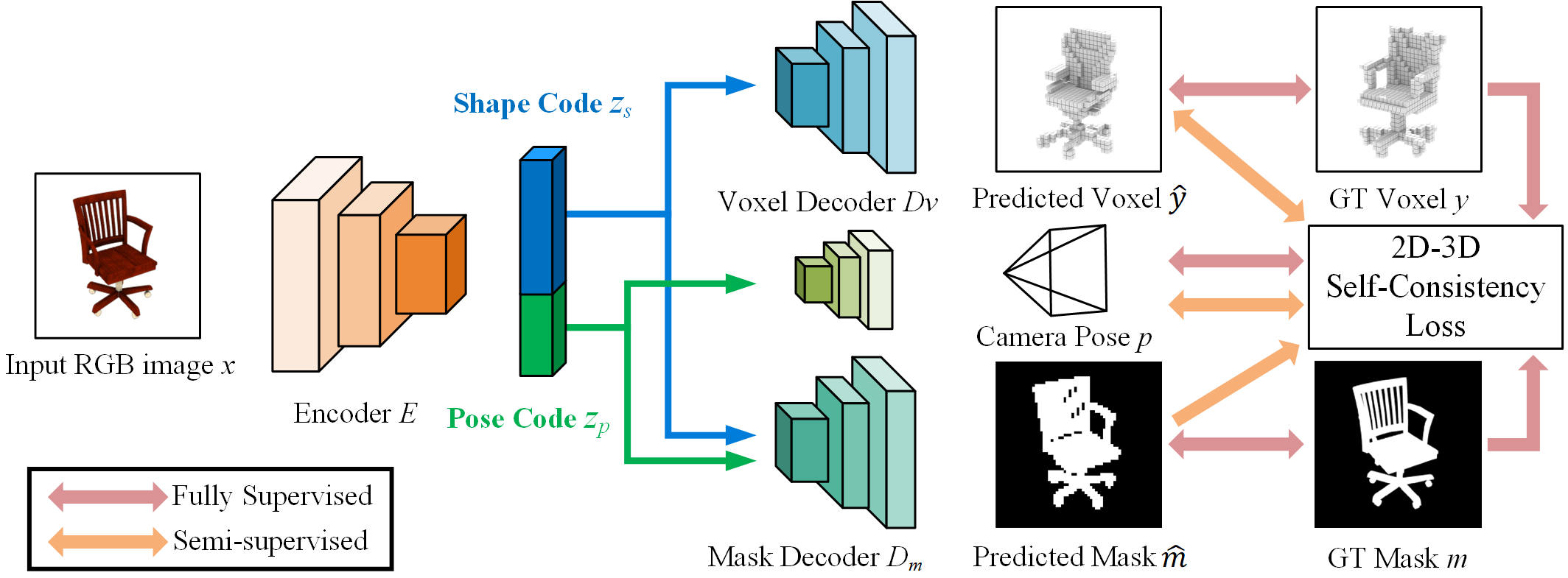}
  \vspace{-2mm}
  \caption{
  Overview of the proposed network architecture, which contains encoder $E$, voxel and mask decoders $D_v$ and $D_m$, respectively. An additional module enforcing 2D-3D self-consistency is introduced, allowing camera pose disentanglement without supervision of its ground truth information.
  }
  \label{fig:sc_training}
\end{figure*}

\section{Pose-Aware 3D Shape Reconstruction via Representation Disentanglement} \label{method} 

In this paper, we propose a unique deep learning framework which observes a single 2D image for 3D shape reconstruction. Our model not only can be extended from fully supervised to semi-supervised settings (i.e., only a portion of 2D images are with ground truth 3D information), it also exhibits abilities in disentangling camera pose information from the derived representation. This disentanglement process is performed in a unsupervised fashion, and is achieved by observing 2D-3D self-consistency as we detail later.

In the remaining of this section, we first describe the notation and architecture in Section \ref{sec:notation_architecture}. How to advance representation disentanglement with 2D-3D self-consistency for 3D reconstruction is detailed in Section \ref{sec:feature_disentangle}. Finally, Section \ref{sec:learning} summarizes learning process of our model. 

\subsection{Notations and Architecture}
\label{sec:notation_architecture}
To reconstruct 3D shape information, we describe shapes in volumetric forms of probabilistic occupancy in this paper. 
For the sake of completeness, we now define the notations which will be used.

Let $\{x_{i}, y_{i}, m_{i}\}^{N}_{i} = \{\textbf{X}, \textbf{Y}, \textbf{M}\}$ denote the training data, 
where $x_i \in \mathbb{R}^{H \times W \times 3}$ indicates the $i^{th}$ input 2D image, $y_i \in \mathbb{R}^{V \times V \times V}$ 
and $m_{i} \in \mathbb{R}^{H \times W}$ are the associated ground truth 3D voxel and 2D mask.
Thus, our goal is to predict the 3D voxel $\hat{y}$ (and the 2D mask $\hat{m}$), while the camera pose information will be jointly predicted in the latent space $z_{p}$.  It is worth noting that, since we focus on a semi-supervised setting, the data used for semi-supervised learning are $\{\textbf{X}\}$ and $\{\textbf{Y}, \textbf{M}\}$, while the size of $\{\textbf{X}\}$ is considered to be larger than that of $\{\textbf{Y}, \textbf{M}\}$.

For the fully supervised version of our framework, every training input image $x$ has the corresponding ground truth 3D voxel $y$ and ground truth 2D mask $m$. And, for the semi-supervised version, only a portion of training images are with their 3D ground truth and 2D masks. Nevertheless, supervised or semi-supervised learning, we never observe ground truth camera pose information during training, while jointly reconstructing 3D shape and camera pose information are our goals.

As shown in Figure~\ref{fig:sc_training}, our proposed network architecture consists of four components:\\

\noindent \textbf{Image Encoder $E$.} The encoder has a residual structure~\cite{resnet}, which maps the input RGB image $x$ into an intrinsic shape representation (i.e., shape code) $z_{s}$ and an extrinsic viewpoint representation (i.e., pose code) $z_{p}$. This disentangled pose code $z_{p}$ is then converted into camera pose $p$ through a FC layer. We note that, translation in the 3D space does not affect the quality of shape reconstruction, and we consider only elevation and azimuth for describing the camera pose.\\

\noindent \textbf{Voxel Decoder $D_v$.} The goal of this decode is to recover the 3D voxels $\hat{y}$ based on the input shape code $z_s$. We follow~\cite{2Ddeconv} for apply 2D deconvolution layers as our voxel tube decoder. However, we choose to apply three separate 2D deconvolution operations, with the channel dimensions of the three deconvolution layers corresponding to height, width, and depth, respectively.\\

\noindent \textbf{Mask Decoder $D_m$.} Different from the voxel decoder $D_v$, the purpose of the mask decoder is to output the 2D mask $\hat{m}$ by observing both the input pose code $z_p$ and shape code $z_s$. We utilize a U-Net~\cite{UNET} based structure for this 2D foreground mask segmentation procedure.\\

\begin{figure}[t]
  \centering	  \includegraphics[width=0.49\textwidth]{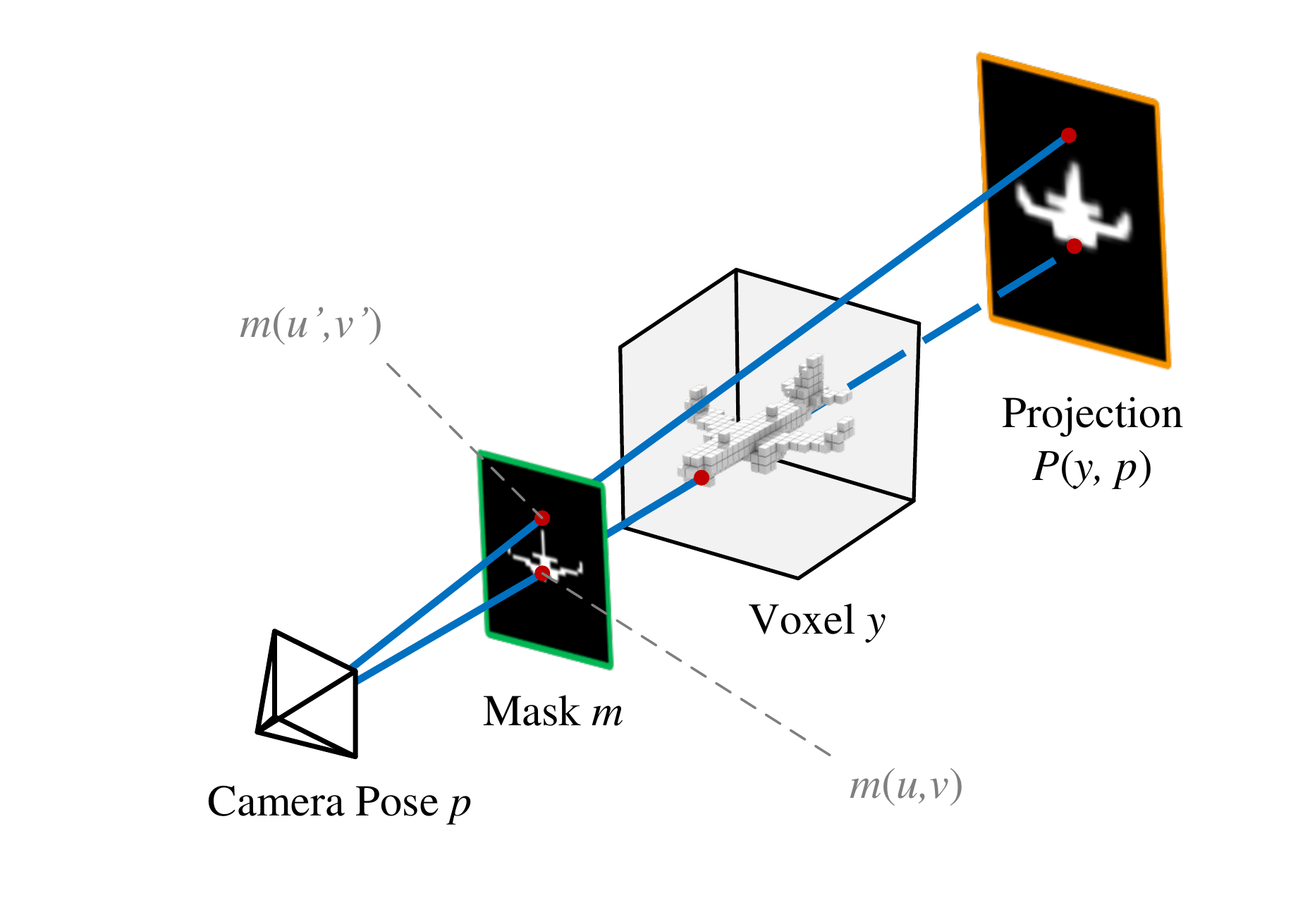}
  \vspace{-8mm}
  \caption{
  Illustration of 2D-3D self-consistency. When the mask value $m$($u$,$v$) is 1, the ray passing through $m$($u$,$v$) would stop at the associated voxel and cannot reach the 2D projection. In contrast, if $m$($u'$,$v'$) is 0, the ray passing through $m$($u'$,$v'$) penetrates the corresponding voxels, and the projected output can observed.
  }
  \label{fig:proj}
\end{figure}

\noindent \textbf{Module for 2D-3D Self-Consistency.}
As a unique design in our network, this module observes the output 3D model and the 2D mask. More precisely, by taking the disentangled camera pose information from the pose code $z_p$, this module enforces the consistency between the predicted 3D voxel and the corresponding 2D mask via the projection loss. As discussed in the following subsection, this is how we achieve disentanglement of extrinsic camera pose from intrinsic shape representation without observing any ground truth information. In other words, the introduction of this module realizes our pose-aware 3D shape reconstruction.

\subsection{Feature Disentanglement via 2D-3D Self-Consistency}
\label{sec:feature_disentangle}

Our proposed model is capable of deep feature disentanglement and pose-aware 3D reconstruction. It is worth repeating that, while our model can be train in fully supervised or semi-supervised settings (i.e., observing ground truth voxels and 2D masks), ground truth camera pose information is never required.

As noted in Sect.~\ref{sec:notation_architecture}, our image encoder $E$ maps input images into an intrinsic shape representation (or shape code) $z_s$ and an extrinsic viewpoint information (or pose code) $z_p$. 
The former is fed into to the voxel decoder $D_v$ to recover pose-invariant 3D voxels, while both $z_s$ and $z_p$ are the inputs to the mask decoder $D_m$ for 2D foreground mask segmentation. In order to convert the extracted pose code into an exact camera pose value (such as rotation and elevation), we transform pose code through one FC layer into camera pose and impose a 2D-3D self-consistency loss, aiming to align the predicted 3D voxels and the 2D mask with the disentangled pose code. We note that, as detailed below, the enforcement of 2D-3D self-consistency is the critical component for camera pose disentanglement, pose-aware 3D reconstruction. It is also the key to enable semi-supervised learning for our proposed network.\\

\noindent \textbf{2D-3D Self-Consistency}
\label{sec:2d_3d_sc}
Differentiable ray consistency loss has been utilized in~\cite{mvc18, MultiViewRay}, which evaluates the inconsistency between mask and shape viewed from predicted camera pose. To achieve camera pose disentanglement, we advance the predicted camera pose information and the corresponding voxel outputs to generate a projection 2D mask, and then calculate the difference between the projection 2D mask and predicted ones from $D_{m}$. This allows the alignment between the the 2D projection of 3D voxels using the disentangled camera pose.

To realize the above disentanglment process via camera pose alignment, we particularly consider a unique 2D-3D self-consistency loss which integrates the projection loss and ray consistency loss as follows
\begin{dmath} \label{eq:sc}
\mathcal{L}_{sc} = {\alpha_{1}\mathcal{L}_{ray} + \alpha_{2}\mathcal{L}_{proj}}.
\end{dmath}

For ray consistency loss, we consider a ray passing through a mask at location ($u$,$v$) to be projected, and traveling along the voxel (as illustrated in Fig.~\ref{fig:proj}).
We sample $N$ values along this ray, and the sampled value represents the occupancy at this sample point.
Next, for each sample point, the probability of the ray stops at that point is calculated.
If the mask value at ($u$,$v$) is 0, the probability $q$ that the ray penetrates across sample points is close to 1. On the other hand, if the mask value at ($u$,$v$) is 1, this probability $q$ would be close to 0.

Now we introduce the details of ray consistency loss.
We consider the ray passing through a mask at location ($u$, $v$).
Given camera intrinsic parameters ($f_u$, $f_v$, $u_0$, $v_0$), where ($f_u$, $f_v$) is focal length of camera and ($u_0$, $v_0$) is optical center of camera, we can determine the direction of this ray as ($\frac{u - u_0}{f_u}$, $\frac{v - v_0}{f_v}$, $1$).
We sample $N$ points along this ray, and the location of $i^{th}$ sample point in the camera coordinate frame is $l_i =$ ( ($\frac{u - u_0}{f_{u}}$)$\frac{i}{N}$, ( $\frac{v - v_0}{f_{v}} $)$ \frac{i}{N}$, $\frac{i}{N}$), where $1$ $\leq$ $i$ $\leq$ $N$.

To calculate the probability of occupancy $y^p_i$ at this sample point, given camera rotation matrix $R$ (parameterized by camera pose $p$) and camera translation $t$, we first map the location of the sample point into $R \times$($l_i+t$).
Then we determine the occupancy $y^p_i$ of the $i^{th}$ sample point by trilinear sampling $Tri$, which is shown as below:

\begin{equation}
    y^p_i = Tri( y, R\times(l_i+t) ).
\end{equation}

Next, the probability of the ray passing through the pixel ($u$,$v$) stops at $i^{th}$ sample point can be obtained, which is shown as below:
\begin{equation} \label{eq:ray_stop}
    q_{u, v}^p(i) = y_i^p\prod_{j=1}^{i-1}(1 - y_j^p) \hspace{3mm} \forall i \leq N.
\end{equation}

If the ray does not stop at any sample point, then it will penetrate the whole voxel.
As a result, we can extend~\eqref{eq:ray_stop} to obtained the probability that this ray that escapes the voxel, and we denote this probability as $q_{(u, v), escape}^p$,

\begin{equation}
    \vspace{-1mm}
    q_{(u, v), escape}^p = \prod_{i=1}^{N}(1 - y_i^p).
    \vspace{-1mm}
\end{equation}

With the above observation, we have the ray consistency loss at pixel ($u$, $v$) defined below:
\vspace{-2mm}
\begin{dmath} \label{eq:ray}
    \mathcal{L}_{ray}(u, v) = m(u, v)q_{(u, v), escape}^p \\ +
    (1-m(u, v))\sum_{i=1}^{N}q_{u, v}^p(i),
\end{dmath}
\vspace{-3mm}
where $m(u, v)$ is the value of the mask at location ($u$,$v$).
If $m(u, v)$ is 0, the probability that the ray penetrates the voxel is close to 1. 
On the other hand, when $m(u, v)$ is 1, the probability that the ray terminates at any sample point is near 0.
As a result, we have the differentiable ray consistency loss $\mathcal{L}_{ray}$($y, p, m$) calculated as the mean of $\mathcal{L}_{ray}$($u, v$) over all pixels in $m$.

In addition to ray consistency, we also consider the projection loss $\mathcal{L}_{proj}$ for 2D-3D self-consistency. As shown in~\eqref{eq:ray}, ($1 - q_{(u, v), escape}^p$) represents the pixel ($u$, $v$) of 3D-2D projection $P$($y$, $p$). As a result, we obtain the projection $P$($y$, $p$) after calculation every ray passing through mask $m$, as shown in Fig~\ref{fig:proj}.
That is, if the ray terminates at that voxels, we expect $q_{(u, v), escape}^p$ to be close to 0, and ($1 - q_{(u, v), escape}^p$) would be near $1$ and represents an object pixel in projection $P$($y$, $p$), as depicted in Fig.~\ref{fig:proj}.
To evaluate the 3D reconstruction, we consider the metric of intersection over union (IoU) and impose the IoU loss~\cite{2Ddeconv} between projection and mask as shown below:
\begin{equation} \label{eq:proj}
\mathcal{L}_{proj}(P(y, p), m) = exp(1-\frac{\sum_{i} P_{i} m_{i}}{\sum_{i} P_{i} + m_{i} + P_{i} m_{i}})-1.
\end{equation}
As~\eqref{eq:sc}-\eqref{eq:proj} are differentiable w.r.t. camera pose $p$ and voxel $y$, this loss would guide the training of both camera pose estimation and 3D shape reconstruction.

In our work, we consider $N=64$ and fix $\alpha_{1} = 5$ and $\alpha_{2} = 0.125$ in \eqref{eq:sc}. 
Because some fine structures like the bases of chairs would diminish when mask is of size $32 \times 32$ as used in~\cite{mvc18}, we use mask of size $48 \times 48$.

Finally, we note that the voxel and mask discussed here can be either ground truth or predicted ones. As shown in Fig.~\ref{fig:sc_training}, we use ground truth voxel and mask for 2D-3D self-consistency loss if full-supervised learning is applicable. If semi-supervised learning is of interest, we then consider the predicted voxel and its mask to calculate the 2D-3D self-consistency loss for the unlabeled data.

\subsection{Learning of Our Model}
\label{sec:learning}

\begin{figure*}[t]
  \centering
  \includegraphics[width=0.99\textwidth]{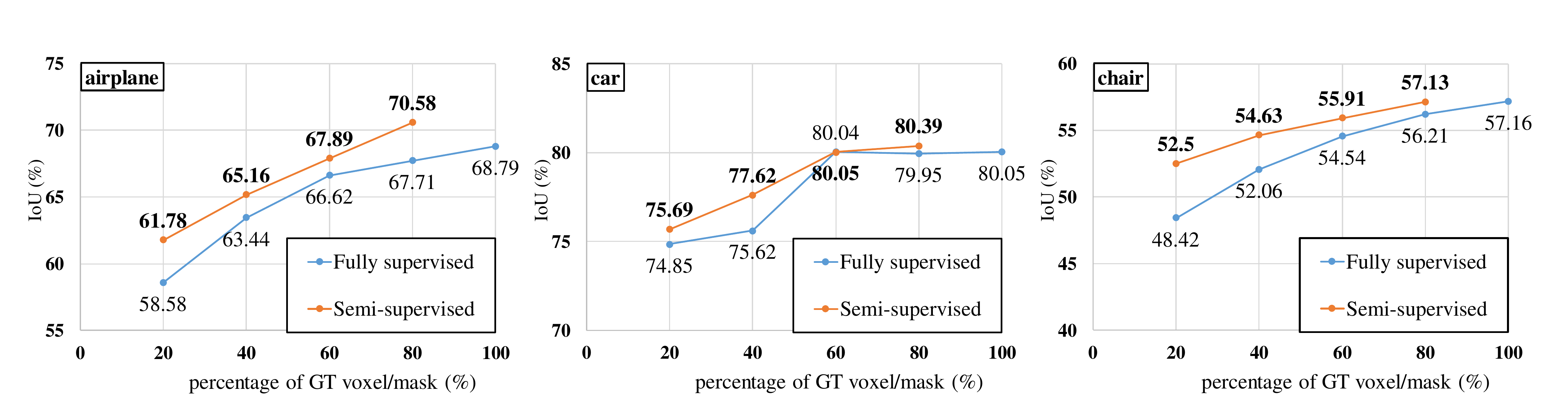}
  \vspace{-3mm}
  \caption{Performance comparisons in IoU using our models training with different degrees of supervision.
  }
  \label{fig:semi_percent}
\end{figure*}

\noindent \textbf{Supervised Learning.}
To train our model in a fully supervised setting, both ground truth voxel $y$ and mask $m$ are observed during training, while the ground truth camera pose is not available. 
The overall loss function for fully supervised learning is shown as below:
\begin{dmath} \label{eq:lossE_stepone}
\mathcal{L}_{sup} = \alpha_{3}\mathcal{L}_{3D} + \alpha_{4}\mathcal{L}_{2D} + \alpha_{5}\mathcal{L}_{sc} + \alpha_{6}\mathcal{L}_{KL}.
\end{dmath}

\noindent Here, we calculate the 2D-3D consistency loss $\mathcal{L}_{sc}$ between the ground truth voxel and mask instead of the predicted ones, which allows the training efficiency and the effectiveness in camera pose disentanglement. 
We use $\mathcal{L}_{3D}, \mathcal{L}_{2D}, \mathcal{L}_{sc}$, and $\mathcal{L}_{KL}$ to update image encoder $E$, and $\mathcal{L}_{3D}$ and $\mathcal{L}_{2D}$ for updating voxel decoder $D_{v}$ and mask decoder $D_{m}$, respectively.
In our work, we fix $\alpha_{3} = 0.5$, $\alpha_{4} = 0.5$, $\alpha_{5} = 1$, and $\alpha_{6} = 0.02$. 

The voxel reconstruction loss $L_{3D}$ consists of positive weighted cross entropy and IoU losses as shown below:
\begin{equation} \label{eq:L3D}
\mathcal{L}_{3D} =  \mathcal{L}_{PCE} + \mathcal{L}_{IoU},
\end{equation}
where $\mathcal{L}_{PCE}$ adopts a positive weight $\alpha_{p}$ to better preserve the fine structures in 3D, and is defined below:
\begin{equation} \label{eq:pwbce}
\mathcal{L}_{PCE} =  -\alpha_{p}\cdot y\cdot log(\hat{y}) - ( 1-y )\cdot log(1-\hat{y}).
\end{equation}
Without this technique, the model tends to predict zeros for the voxels corresponding to such structures which minimizes the overall cross entropy loss. As for the IoU loss $\mathcal{L}_{IoU}$~\cite{2Ddeconv}, it is calculated as:
\begin{equation} \label{eq:iou}
\mathcal{L}_{IoU}( \hat{y}, y) = exp(1-\frac{\sum_{i} \hat{y_{i}} y_{i}}{\sum_{i} \hat{y_{i}} +  y_{i} + \hat{y_{i}} y_{i}})-1.
\end{equation}
For 2D mask segmentation, we calculate the cross entropy between the predicted and ground truth masks as our 2D reconstruction loss $L_{2D}$. 

As for the KL divergence loss, it is enforced to regularize the distribution of shape code $z_{s}$ and model ambiguity of 3D reconstruction due to unseen parts of shapes~\cite{pointset}. We adopt conditional variational autoencoder~\cite{AAE, VAE} for $E$ and $D_v$ as shown below:
\begin{equation} \label{eq:kl}
\mathcal{L}_{KL} = \mathcal{KL}(\mathcal{N}(z_{s, u}, z_{s, var}) \| \mathcal{N}(0, 1)),
\end{equation}
 
\noindent where the shape code $z_s$ consists of mean $z_{s, mu}$ and variance $z_{s, var}$. Only the mean $z_{s, mu}$ is utilized by the mask decoder $D_m$ as mask segmentation is under-determined.\\

\noindent \textbf{Semi-supervised Learning.}
In practice, we cannot collect the ground truth 3D voxels and 2D masks for all images, and thus a semi-supervised setting would be of interest. In this setting, only a portion of input images $x$ are with ground truth $y$ and $m$, while the remaining data for training are the 2D images $x$ only. With such training data, we first pre-train our network using fully supervised data, followed by fine-tuning the network using unlabeled input images only. To be more specific, for semi-supervised learning, the overall loss function is defined as below:
\begin{dmath} \label{eq:lossE_semi}
\mathcal{L}_{semi} = \mathcal{L}_{sc} + \alpha_{6}\mathcal{L}_{KL}.
\end{dmath}
Note that $\mathcal{L}_{sc}$ and $\mathcal{L}_{KL}$ are calculated to update image encoder $E$, while $\mathcal{L}_{sc}$ for updating voxel decoder $D_{v}$. 

We note that, in the network refinement process under this semi-supervised setting, our introduced 2D-3D self-consistency is critical. This allows our model to align the predicted 3D voxel $\hat{y}$ and the predicted 2D mask $\hat{m}$ with disentangled camera pose $p$, all in an unsupervised fashion.

\begin{table}[t]
    \begin{tabular}{|L{2.5cm}|| C{0.9cm}  C{0.9cm}  C{0.9cm} C{0.9cm}|}
		\hline
        & \multirow{2}{*}{airplane} & \multirow{2}{*}{car} & \multirow{2}{*}{chair} & \multirow{2}{*}{Mean}\\
        Method   &       &          &     & \\
        \hline
		3D-R2N2 \cite{3DR2N2} & 51.3 & 79.8 & 46.6 & 59.2 \\
		OGN~\cite{OGN} & 58.7 & 81.6 & 48.3 & 62.9 \\
		PSGN \cite{pointset} & 60.1 & 83.1 & 54.4 & 65.9 \\
		voxel tube \cite{2Ddeconv} & 67.1 & 82.1 & 55.0 & 68.1 \\
		Matryoshka \cite{2Ddeconv} & 64.7 & 85.0 & 54.7 & 68.1 \\
		Ours & \textbf{69.2} & \textbf{85.8} & \textbf{56.7} & \textbf{70.6} \\
        \hline
	\end{tabular}
    \caption{Comparisons of 3D shape reconstruction in fully supervised settings in terms of mean IoU (\%).}
	\label{table:fully_iou}
\end{table}

\section{Experiments}
\subsection{Dataset}
We consider the ShapeNet dataset~\cite{ShapeNet} which contains a rich collection of 3D CAD models, and is widely used in recent research works related to 2D/3D data.
Three categories, airplane, car, and chair, are selected for our experiments. For fair comparisons, we consider two different data settings. For supervised learning of our model and to perform comparisons, we follow the works of 3D-R2N2~\cite{3DR2N2}, Octree Generating Network (OGN)~\cite{OGN}, Point Set Generation Network (PSGN) \cite{pointset}, voxel tube network and Matryoshka network~\cite{2Ddeconv}, which scale the ground truth voxels to fit into $32\times32\times32$ grids. This makes ground truth voxels larger than those considered in MVC~\cite{mvc18} and DRC~\cite{MultiViewRay}. We use the same rendered images, ground truth voxel, and data split as used in these work.

For semi-supervised learning, we generate $24$ rendered images of size $64 \times 64 \times 3$ pixels and the corresponding ground truth 2D masks, using the same camera pose information and data split as used in Perspective Transformer Nets (PTN)~\cite{NIPS2016_6206}. We utilize the same ground truth voxels in Multi-view Consistency (MVC)~\cite{mvc18} and Differentiable Ray Consistency (DRC)~\cite{MultiViewRay} to fit our projection module, and the grid size of voxels is $32\times32\times32$.

\subsection{Fully Supervised Learning} \label{sec:exp_ful}
To train our model using fully-supervised data, we consider state-of-the-art methods of~\cite{3DR2N2, OGN, pointset, 2Ddeconv} for comparisons. Since both ground truth voxels and 2D masks are available during training, there is no need to utilize the 2D-3D self-consistency loss for camera pose disentanglement.

Quantitative results of different 3D reconstruction methods are listed in Table.~\ref{table:fully_iou}. From this table, we see that our network achieved favorable performances, showing that our network architecture is preferable under such settings.

\begin{table*}[t]
	\centering
    \begin{tabular}{|L{1.8cm}|| C{2.6cm} C{2.6cm} C{2.6cm} || C{1.0cm} C{1.0cm} C{1.0cm} C{1.0cm}| }
		\hline
        &  \multirow{2}{*}{Training data} & \multirow{2}{*}{Label percentage} & \multirow{2}{*}{Single/multi views} & \multirow{2}{*}{airplane} & \multirow{2}{*}{car} & \multirow{2}{*}{chair} & \multirow{2}{*}{Mean} \\
        Method & & & & & & & \\
        \hline
		DRC \cite{MultiViewRay} & 2D mask \& pose & 100\% & M & 53.5 & 65.7 & 49.3 & 56.2 \\
		MVC \cite{mvc18} & 2D mask &  100\% & M & 49.4 & 63.9 & 39.6 & 51.0 \\
		Ours (Semi) & 2D mask \& voxel& 15\% & S & 59.6 & 75.3 & 52.3 & 62.4\\
		Ours (Sup) & 2D mask \& voxel &  100\% & S & 68.8 & 80.1 & 57.2 & 68.7 \\
        \hline
	\end{tabular}
    \caption{Comparisons in IoU with approaches not requiring ground truth 3D shape information in weakly or semi-supervised settings. Note that Ours (Sup) denotes the fully-supervised version of our method.}
	\label{table:sup_comparison}
\end{table*}

\subsection{Semi-Supervised Learning} \label{sec:exp_semi}
%

To demonstrate the effect of our semi-supervised learning, we use different portions of labeled/unlabeled images to train the network for 3D reconstruction. For comparison purposes, we use 100, 80, 60, 40, 20$\%$ of labeled images for training fully-supervised version of our model, and the rest of the images are unused. As for semi-supervised learning, we use 80, 60, 40, 20$\%$ of labeled images, plus the remaining unlabeled ones for training our model. 

\begin{figure}[t]
  \centering	  
  \includegraphics[width=0.495\textwidth]{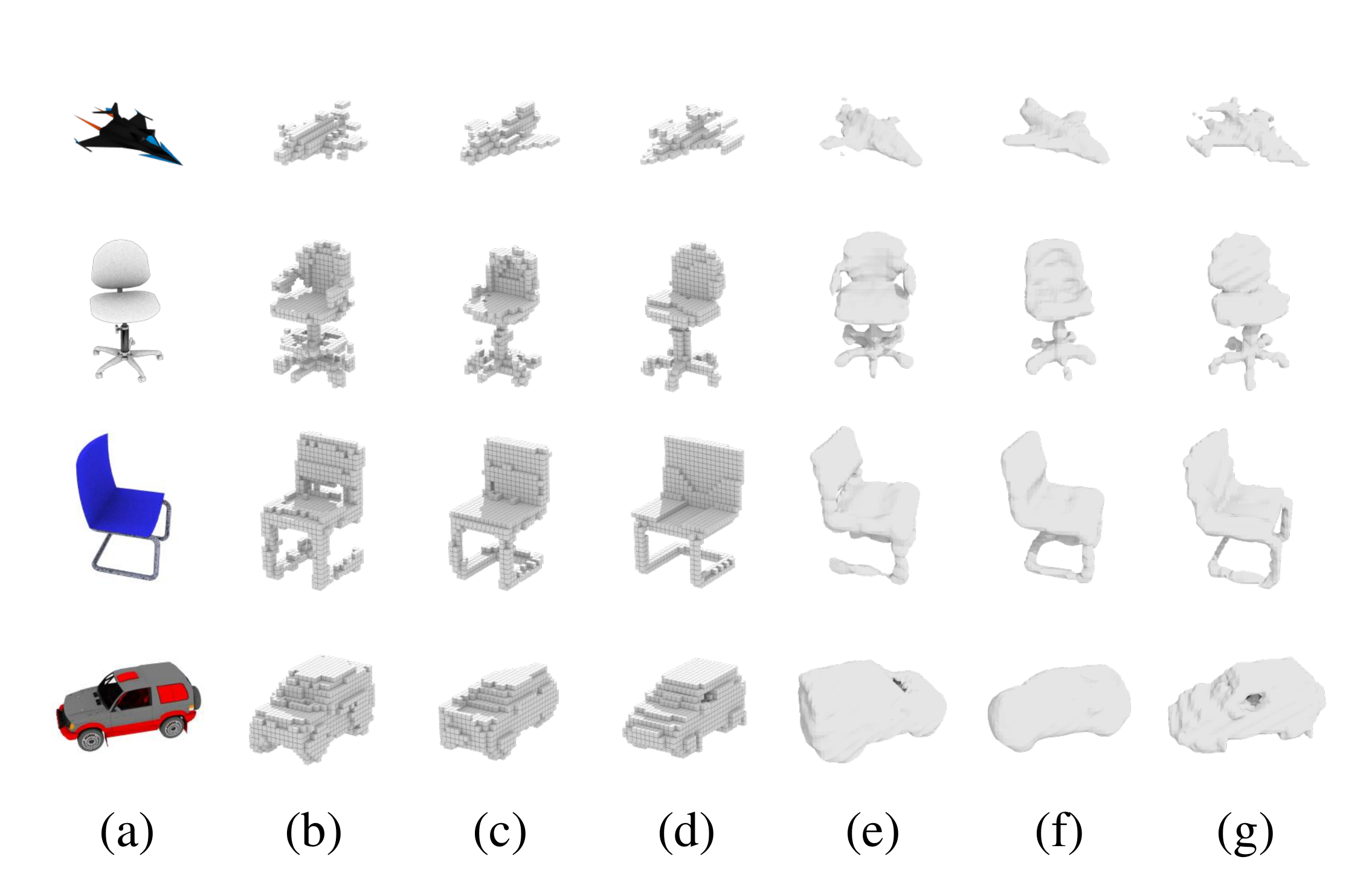}
  \vspace{-4mm}
  \caption{Visualization of shape reconstruction using our model. (a) Input image $x$, (b) predicted voxel $\hat{y}$ using our supervised version, (c) predicted voxel $\hat{y}$ using our semi-supervised model, (d) ground truth voxel $y$, (e) mesh of (b), (f) mesh of (c), and (g) mesh of (d). Note that (e)-(g) are produced using the disentangled camera pose information.}
  \vspace{-3mm}
  \label{fig:semi_shape}
\end{figure}

We compare the results of our models with fully supervised and semi-supervised learning, and compare the quantitative results in Fig.~\ref{fig:semi_percent}. From this table, we see that our semi-supervised version exhibited very promising capabilities in handling unlabeled data, which resulted in improved performances compared to the versions using only labeled data for training. Qualitative results are also shown in Fig.~\ref{fig:semi_shape} for visualization purposes.

\begin{figure}[t]
  \centering	  
  \includegraphics[width=0.495\textwidth]{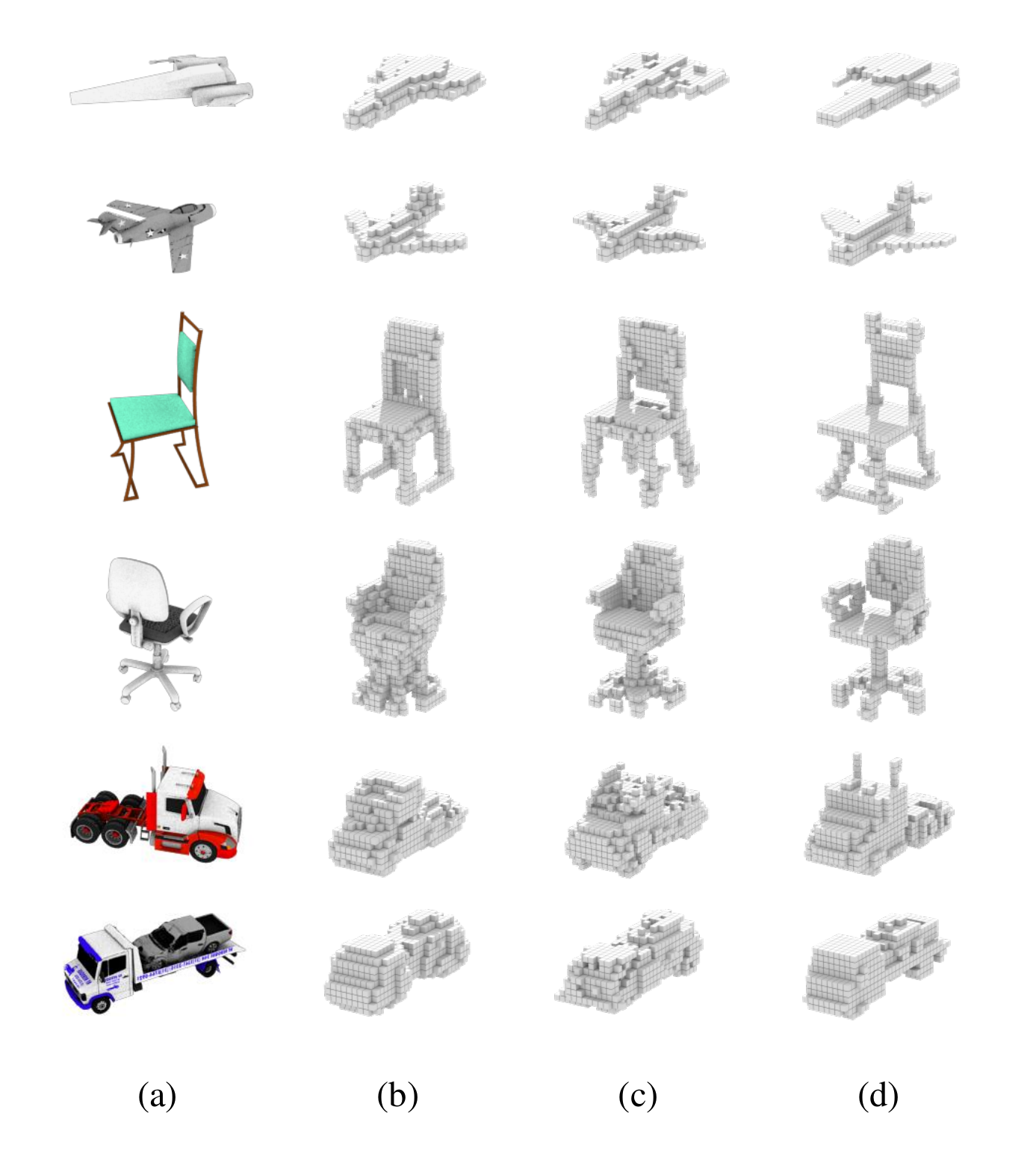}
  \vspace{-8mm}
  \caption{Visualization of shape reconstruction using different weakly or semi-supervised methods. (a) Input image $x$, (b) predicted voxel of DRC~\cite{MultiViewRay}, (c) predicted voxel of our method, and (d) ground truth voxel $y$. Note that DRC requires supervision of pose information, while ours does not.}
  \vspace{-3mm}
  \label{fig:weakly_shape}
\end{figure}

To perform more complete comparisons with recent approaches not requiring ground truth 3D shape information, we consider the works of MVC~\cite{mvc18} and DRC~\cite{MultiViewRay}. Note that both DRC and MVC require multi-view images as inputs for training; morevoer, DRC requires ground truth camera pose information. Table.~\ref{table:sup_comparison} lists and compares the performances, in which our method clearly achieved the best results among all. Fig.~\ref{fig:weakly_shape} additionally presents qualitative results, which visually present and compare the 3D reconstruction abilities of different models.

\subsection{Camera Pose Prediction}
\begin{figure}[t]
  \centering	  
  \includegraphics[width=0.49\textwidth]{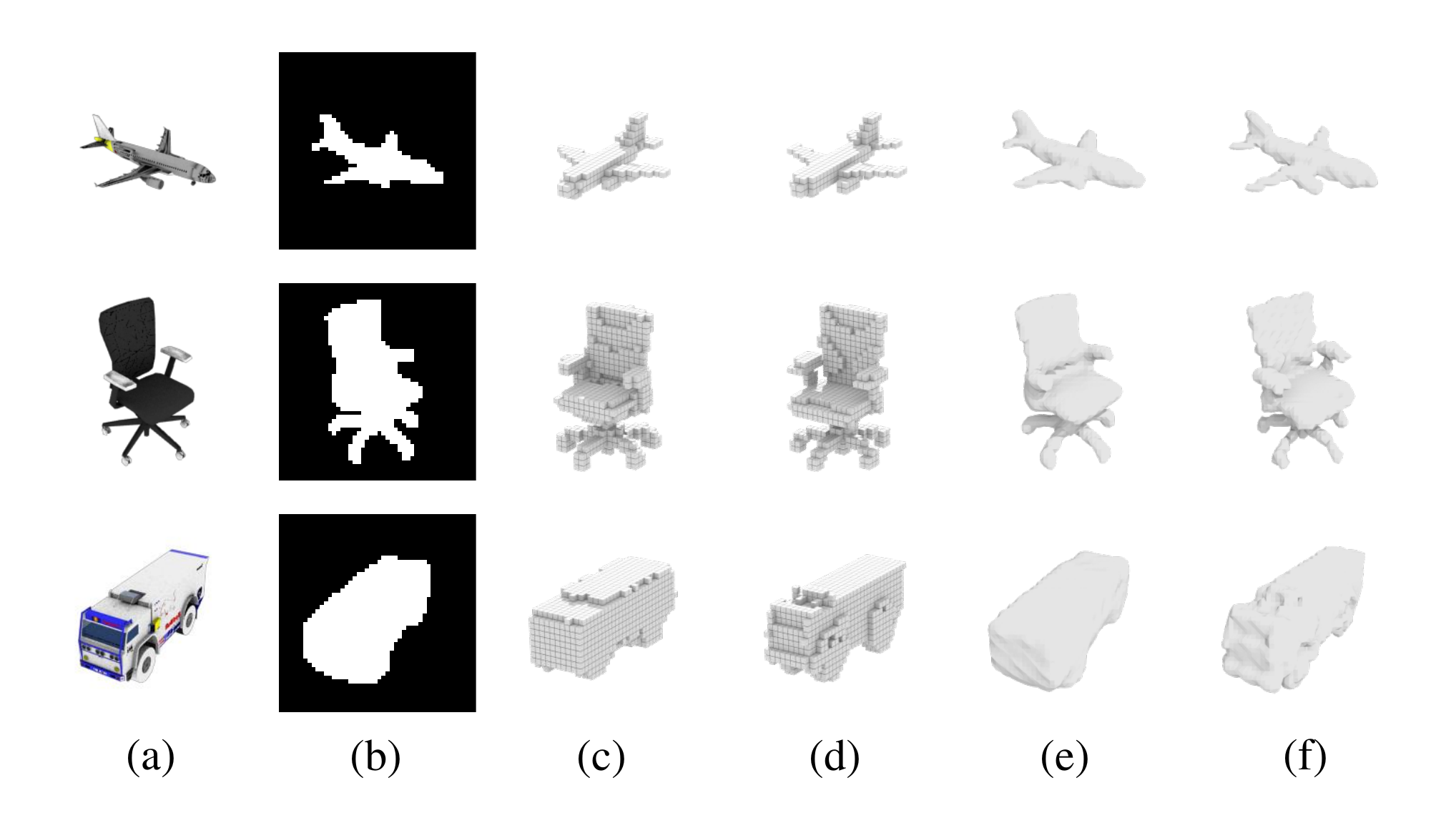}
  \vspace{-4mm}
  \caption{
  Visualization of pose-aware 3D reconstruction. (a) Input image $x$, (b) predicted 2D mask, (c) predicted 3D voxels using our disentangled pose, and (d) ground truth voxel $y$, (e) mesh of (c), and (f) mesh of (d).}
  \label{fig:shape}
\end{figure}

To demonstrate the ability of our model in disentangling camera pose information without such supervision, we transform our voxels into pose-aware shapes with our predicted and manipulated camera poses, as depicted in Fig.~\ref{fig:shape}.
From the results shown in figure, we see that the use of our model for learning interpretable visual representations from 2D images, including shape and camera pose features, can be successfully verified.

We also present the quantitative results of camera pose prediction compared to MVC~\cite{mvc18} in Table.~\ref{table:camera_pose_pred}. It is worth pointing out that, MVC~\cite{mvc18} requires multi-view images as their inputs. From this table, we see that our model succesfully disentangle and predict the camera pose information, without supervision of such information.

\subsection{Ablation Studies} \label{sec:ablation}
To assess the contributions of each component and the design of our proposed network architecture, we perform the following ablation studies.\\

\noindent \textbf{Network Architecture.}
Recall that our proposed voxel decoder consists of three separate 2D deconvolution layers in three directions, with each of the channel dimensions corresponding to height, width, and depth, respectively. We compare our proposed voxel decoder with the one using 2D deconvolution layers in only one direction~\cite{2Ddeconv}.
For fair comparisons, we fix other components in our network, and compare the performances of the networks utilizing these two different kinds of voxel decoders. From the results shown in Table~\ref{table:voxtube_comparison}, we see that our decoder design resulted in improved 3D shape reconstruction performances.\\

\noindent \textbf{2D-3D Self-Consistency.}
We assess the contribution of our network module in observing 2D-3D self-consistency. For simplicity, we only consider ray consistency loss with and without the projection loss, and we use ground truth camera pose to calculate~\eqref{eq:ray} and~\eqref{eq:proj}. Then, we compare the resulting 3D reconstruction results.

From the comparison results shown in Table~\ref{table:proj_comparison}, we confirm that the introduced 2D-3D consistency combining both ray consistency and projection losses would improve the performance of 3D reconstruction. Thus, exploiting this property for 3D shape reconstruction would be preferable.

\begin{table}[t]
	\centering
    \begin{tabular}{|l||cc|}
		\hline
        Method &  airplane & chair\\
        \hline
        MVC~\cite{mvc18} & 35.10\degree & 9.41\degree\\
        Ours (Sup) & 11.24\degree & 9.82\degree\\
        Ours (Semi)  & 9.34\degree &  6.24\degree\\
        \hline
	\end{tabular}
    \caption{Mean prediction error of camera poses. Note that MVC requires multiple input images, while supervised/semi-supervised versions of our model always observe a single input without utilizing ground truth pose information. Note that Ours (Sup) is trained using only 15\% of the data with ground truth shape/mask information, while Ours (Semi) utilizes both 15\% labeled and 85\% unlabeled data for training.}
	\label{table:camera_pose_pred}
\end{table}

\section{Conclusion}

In this paper, we proposed a deep learning framework for single-image 3D reconstruction.
Our proposed model is able to learn deep representation from a single 2D input for recovering its 3D voxel.
This is achieved by disentangling unknown camera pose information from the above features via exploiting 2D-3D self-consistency. More importantly, no camera pose information, classification labels, or discriminator is required.
Our method can be trained on fully-supervised setting as most 3D shape reconstruction models do, which utilize 2D images and their ground truth 3D voxels for training.
Also, it can be trained on semi-supervised settings, which use additional unlabeled 2D images to further enhance the reconstruction results.
Both quantitative and qualitative results demonstrated that our method was able to produce satisfactory results when comparing to state-of-the-art approaches with fully or semi-supervised settings. Thus, the effectiveness and robustness of our model can be successively verified.

\begin{table}[t]
	\centering
    \begin{tabular}{|L{2.5cm}|| C{0.875cm} C{0.875cm} C{0.875cm} C{0.875cm} |}
		\hline
        \multirow{2}{*}{Decoder design}& \multirow{2}{*}{airplane} & \multirow{2}{*}{car} & \multirow{2}{*}{chair} & \multirow{2}{*}{Mean}\\
        & & & & \\
        \hline
		1 deconv & 68.6 & 83.7 & 56.2 & 69.5 \\
		3 deconvs & \textbf{69.2} & \textbf{85.8} & \textbf{56.7} & \textbf{70.6} \\
        \hline
	\end{tabular}
    \caption{Comparisons of voxel decoder designs. Note that \textit{1 deconv} indicates the voxel decoder using 2D deconvolution layers in only one direction, while \textit{3 deconvs} denotes our design using doconvolution layers in three directions.
    }
	\label{table:voxtube_comparison}
\end{table}

\begin{table}[t]
	\centering
    \begin{tabular}{|L{2.5cm}|| C{0.875cm} C{0.875cm} C{0.875cm} C{0.875cm} |}
		\hline
        \multirow{2}{*}{Method} & \multirow{2}{*}{airplane} & \multirow{2}{*}{car} & \multirow{2}{*}{chair} & \multirow{2}{*}{Mean}\\
        & & & & \\
        \hline
		Ray & 53.5 & 65.7 & 49.3 & 56.2 \\
		Ray + Projection  & \textbf{55.7} & \textbf{66.0} & \textbf{49.6} & \textbf{57.1} \\
        \hline
	\end{tabular}
    \caption{Evaluation of different losses for 2D-3D self-consistency. \textit{Ray} means that only use ray consistency loss, and \textit{Ray + Projection} means using both ray consistency loss and projection loss.
    }
	\label{table:proj_comparison}
\end{table}

{\small
\bibliographystyle{ieee}
\bibliography{egbib}
}

\newpage

\section*{Appendix}
\begin{appendices}

\section{Implementation Details} \label{sec:imp}
We implement the proposed network by PyTorch. The image encoder $E$, voxel decoder $D_{v}$, and mask decoder $D_{m}$ are not pre-trained and are all randomly initialized.
We choose to use ADAM optimizer to train $E$, $D_{v}$ and $D_{m}$.
With fully-supervised learning settings, the learning rate for $E$, $D_{v}$, and $D_{m}$ is set to $2\times10^{-4}$. 
As for the semi-supervised learning, the learning rates for $E$, $D_{v}$, and $D_{m}$ are all set to $2\times10^{-5}$.
The batch size is set to 48.
The positive weight $\alpha_{p}$ for $L_{PCE}$ is set to 3.
We train our model on a single NVIDIA GeForce GTX 1080 Ti GPU with 11 GB memory. More details about the network architecture is in Sec.~\ref{sec:net_archi}. Besides, we will release our code of this work so that more details can be shown.

\section{Network Architecture} \label{sec:net_archi}
We now describe the detailed network architectures, including image encoder $E$, mask decoder $D_m$, and voxel decoder $D_v$. \\

\noindent \textbf{Image Encoder $E$.}
Our image encoder $E$ has residual structures. More specifically, it is composed of five designed residual blocks.
%
The activation function for each block output is a leaky rectified linear unit (Leaky ReLU).\\

\begin{figure}[t]
  \centering	  
  \includegraphics[width=0.495\textwidth]{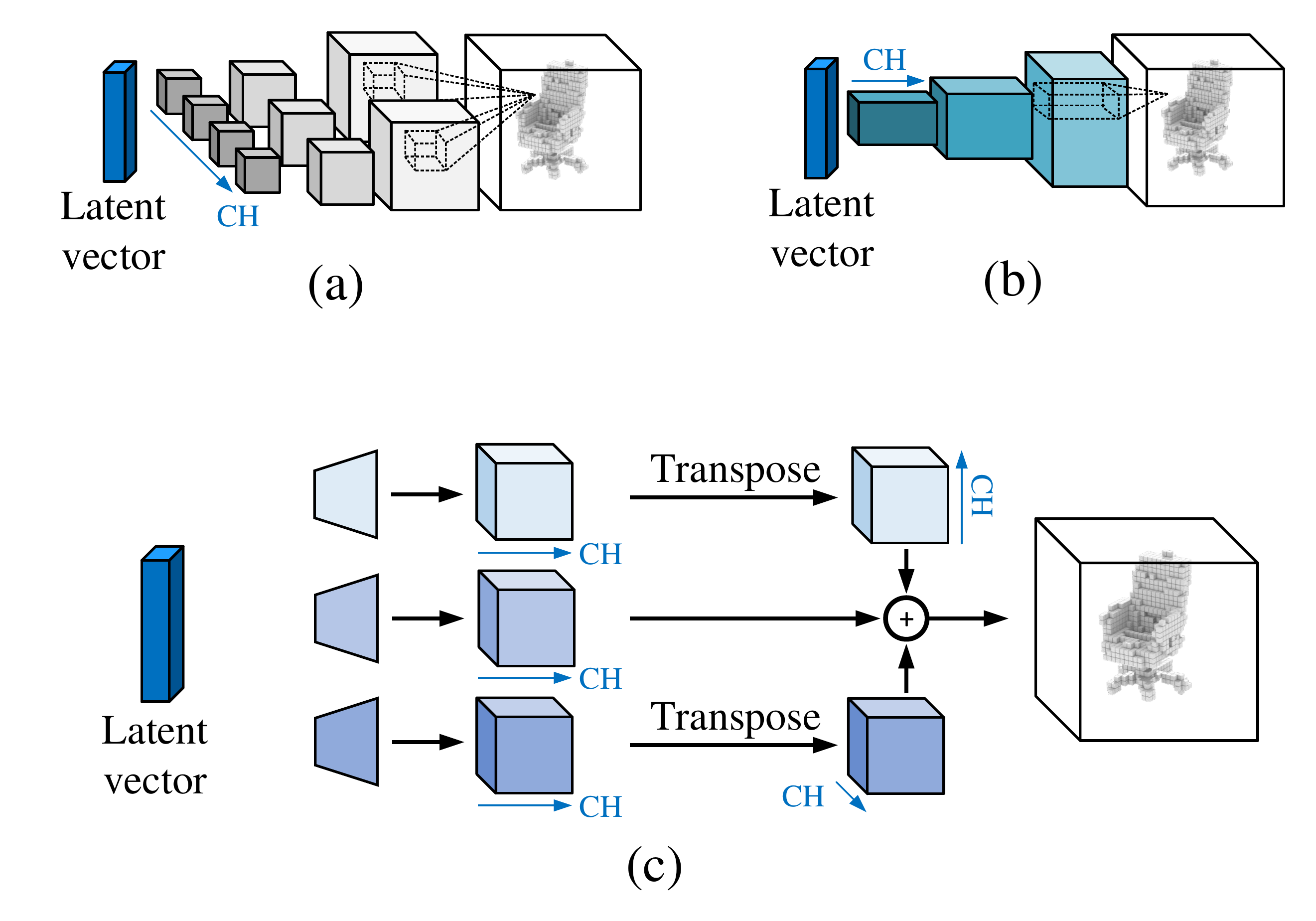}
  \caption{Different voxel decoder designs. (a) 3D deconvolution, (b) one 2D deconvolution, and (c) our proposed module with three 2D deconvolutions. Note that while the uses of 2D deconvlutions require fewer number of network parameters than that of 3D deconvolutions, our design using three 3D deconvolution achieves the best performance than those using either one 2D or full 3D deconvolutions.
  }
  \label{fig:dec_v}
\end{figure}

\noindent \textbf{Mask Decoder $D_{m}$.}
Our mask decoder $D_{m}$ has a U-Net based structure.
It is composed of 5 upsampling blocks.\\

\begin{figure*}[t]
  \centering	  
  \includegraphics[width=0.95\textwidth]{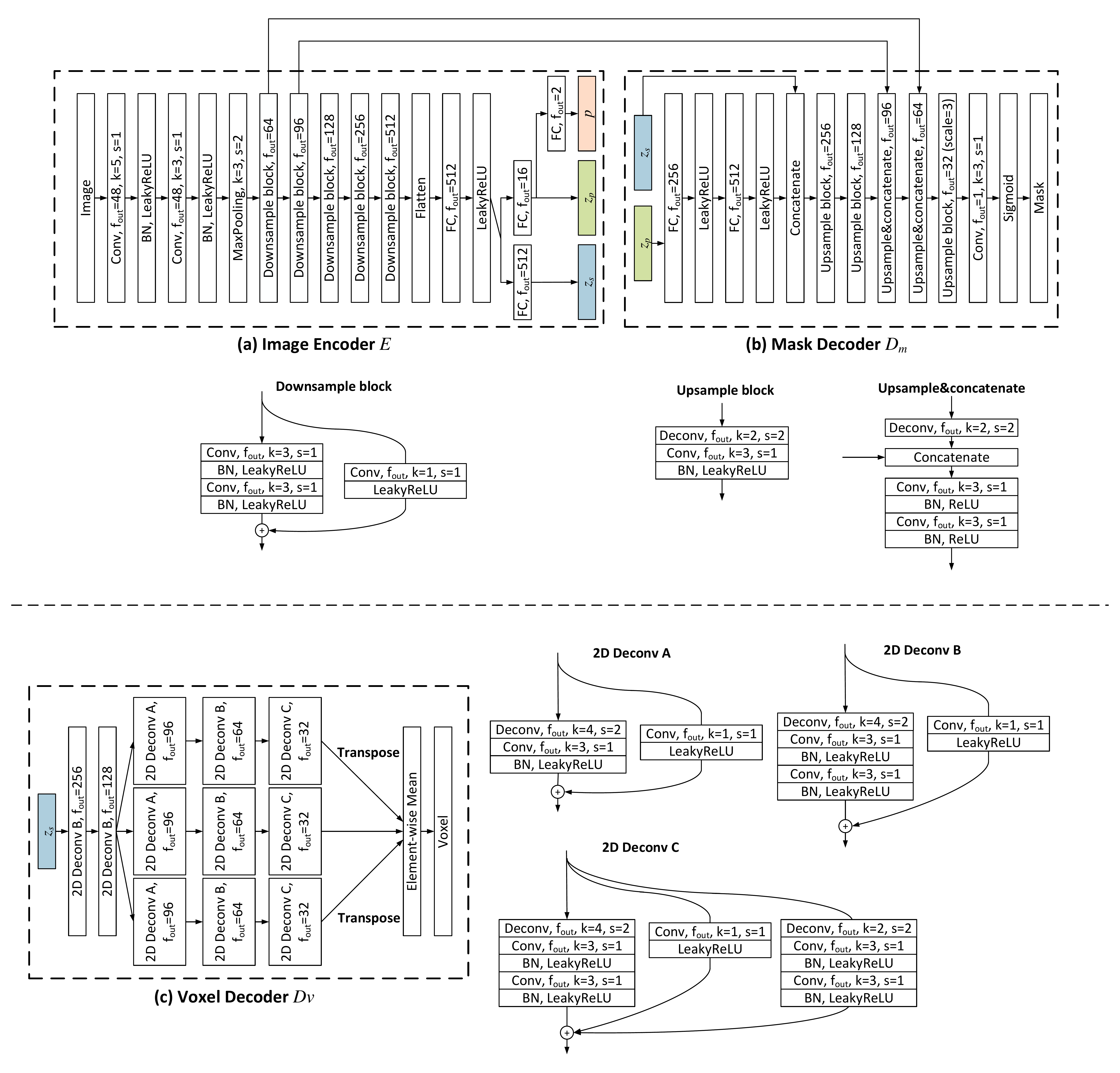}
  \caption{Detailed designs of each component in our network architecture. 
  (a) Image encoder $E$, (b) mask decoder $D_{m}$, and (c) voxel decoder $D_{v}$.
  }
  \vspace{-3mm}
  \label{fig:model}
\end{figure*}

\noindent \textbf{Voxel Decoder $D_{v}$.}
The proposed voxel decoder contains three separate 2D deconvolution layers, each of the three channel dimensions corresponding to height, width, and depth, respectively.
As illustrated in Fig.~\ref{fig:dec_v}, the two output tensors are first transposed so that the channels would be mapped to different spatial dimensions. 
We take the element-wise mean from the three tensors to obtain the predicted shape.
The complete architecture design of each network component is shown in Fig.~\ref{fig:model}.\\



\section{Visualization and Comparisons}

\begin{figure*}[t]
  \centering	  
  \includegraphics[width=0.895\textwidth]{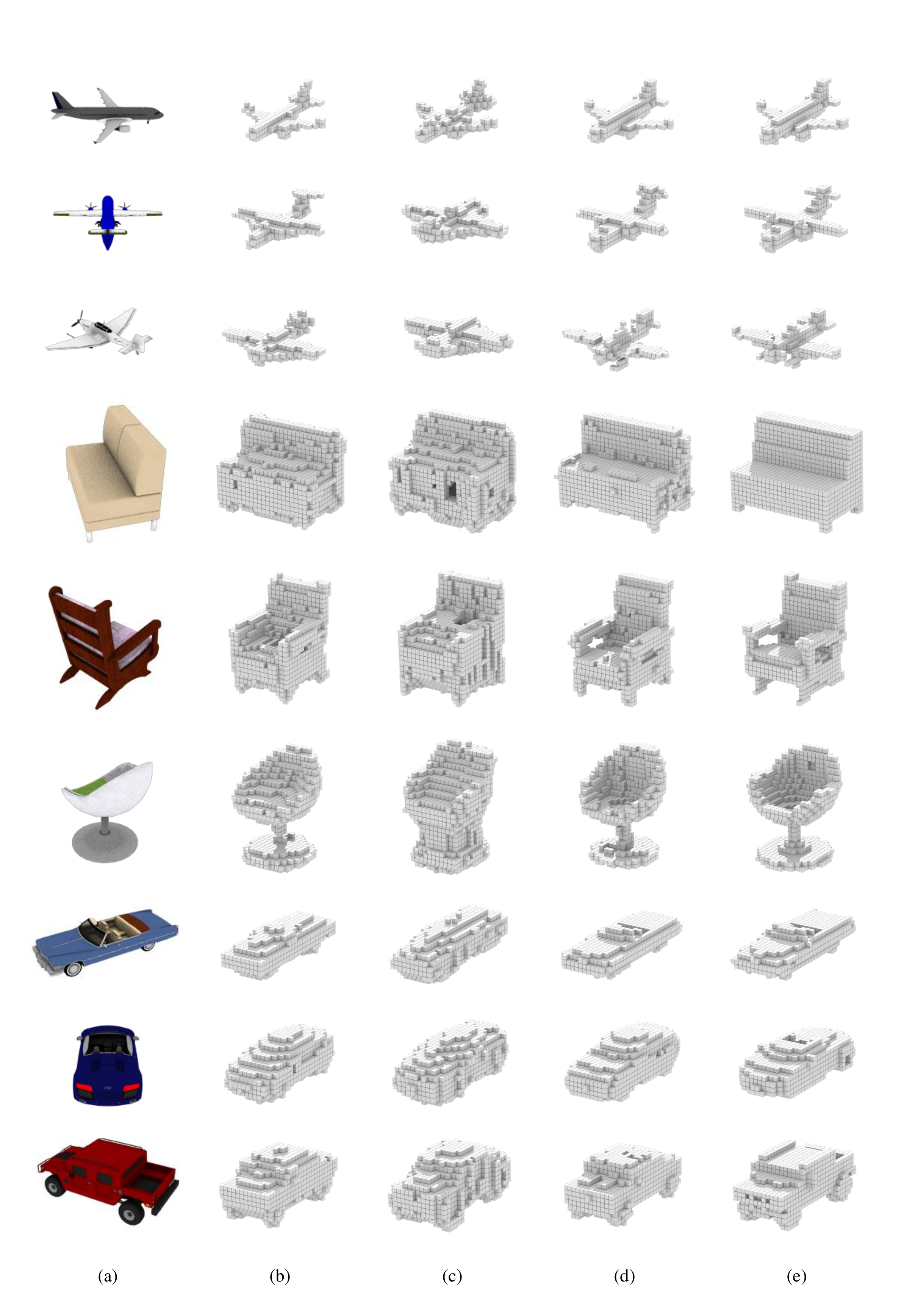}
  \vspace{-8mm}
  \caption{
  Visualization  of  shape  reconstruction  using  different weakly or semi-supervised methods.
  Note that our semi-supervised method utilized both 15\% labeled and 85\% unlabeled data for training.
  (a) Input image, (b) predicted voxel of DRC, (c) predicted voxel of MVC, (d) predicted voxel of our method, and (e) ground truth voxel.
  }
  \vspace{-3mm}
  \label{fig:comp}
\end{figure*}

We now provide additional visualization of 3D shape reconstruction using different weakly or semi-supervised methods, including ours. Note that both DRC and MVC require multi-view images as inputs for training. In addition, DRC requires ground truth camera pose information when learning its model. Example results are shown in Fig.~\ref{fig:comp}.

\end{appendices}

\end{document}